\documentclass[runningheads]{llncs}
\usepackage[T1]{fontenc}
\usepackage{graphicx}
\usepackage{verbatim}

\begin{document}
\title{Effects of Soft-Domain Transfer and Named Entity Information on Deception Detection}
\titlerunning{Effects of Soft-Domain Transfer on Deception Detection}
%
\author{Steven Triplett\inst{1} \and Simon Minami\inst{2} \and Rakesh Verma\inst{1}\orcidID{0000-0002-7466-7823} }
%
%
\institute{University of Houston, Houston, TX, USA \\
\email{smtriple@CougarNet.UH.EDU}
\email{rmverma2@Central.UH.EDU}
\and Tufts University\\
\email{simon.sho.minami@gmail.com}}
\maketitle              
\begin{abstract}
 In the modern age an enormous amount of communication occurs online, and it is difficult to know when something written is genuine or deceitful. There are many reasons for someone to deceive online (e.g., monetary gain, political gain) and detecting this behavior without any physical interaction is a difficult task. Additionally, deception occurs in several text-only domains and it is unclear if these various sources can be leveraged to improve detection. To address this, eight datasets were utilized from various domains to evaluate their effect on classifier performance when combined with transfer learning via intermediate layer concatenation of fine-tuned BERT models. We find improvements in accuracy over the baseline. Furthermore, we evaluate multiple distance measurements between datasets and find that Jensen-Shannon distance correlates moderately with transfer learning performance. Finally, the impact was evaluated of multiple methods, which produce additional information in a dataset's text via named entities, on BERT performance and we find notable improvement in accuracy of up to 11.2\%.
\keywords{Deception \and BERT \and Fake News \and Transfer learning}
\end{abstract}
\section{Introduction}
In the modern age an enormous amount of communication occurs online through the internet. This communication can impact the opinions of others through the posting of reviews or through social media networks. Moreover, as more commerce is conducted through the internet, the incentive to deceive others by misconstruing the quality of a product is growing. One study found that 80\% of consumers changed their original decision to purchase a commodity due to the presence of many negative reviews for that commodity, and 87\% of consumers will decide to purchase a commodity based upon positive reviews \cite{tang2020review}. Another medium for deception is through news and journalism. The rise of fake and deceptive news reporting has created definitive impacts on society and led to a tumultuous political ecosystem \cite{lazer2018science}. 

Deception has been defined as a deliberate attempt to misconstrue or create a false impression~\cite{burgoon1994interpersonal} and deception detection is known to be a challenging task~\cite{fornaciari2021bertective,verma2024}. Researchers in experimental psychology have conducted studies over nearly a century to investigate different cues that were assumed to reveal deception, but despite these efforts, the results show that humans are simply not good at determining truth from lies \cite{hauch2015computers}. Machine learning (ML) approaches utilizing techniques from Natural Language Processing (NLP) have proven successful, but the success is predominantly in areas such as online reviews~\cite{li2021exploring,wang2012identify,liu2020detecting} and phishing emails~\cite{egozi2018phishing,aasalBD2020}. Certain feature spaces that prove successful in one deceptive domain may not necessarily guarantee success in another. Therefore, an open question is to develop an automated ML-based classification framework for deception detection of text that can utilize data from other similar deceptive domains to improve performance. 

To address the above goal in this study, we utilized eight quality datasets from various sources to evaluate their effect on classifier performance when transfer learning is leveraged. We select a balanced deception dataset as the target dataset and experiment with multiple transfer learning methods including boosting-based methods and a deep learning based soft-domain transfer method.  Additionally, we alter these datasets using multiple methods of manipulating information based on named entities in the text and examine the impact this has on classification. We also study whether distance functions can explain the performance of transfer learning methods and whether the size of the source dataset versus the target dataset size can have an impact on performance. To summarize, we make the following contributions:

\begin{enumerate}
    \item We show that for deception detection, the deep learning based transfer method outperforms boosting-based transfer learning methods.  
    \item We find that the source versus target relative dataset size is a significant factor in transfer learning performance for deception detection. 
    \item We find that significant increases in accuracy of deception detection when compared to baseline performance are achieved when replacing named entities (NE) with explanations, replacing named entities with the respective part-of-speech (POS) tag, and when attaching the explanation to the named entity. The cumulative effect of transfer learning and named-entity processing is substantially higher than the base model directly trained on the target NE-unprocessed dataset. 
    \item We show that the Jensen-Shannon Divergence and the source to target KL-divergence have moderate correlations (Spearman rho) with the performance of transfer-learning methods. 
\end{enumerate}

The rest of the paper is organized as follows. In the next section, we discuss the related work. In Section~\ref{sec:data}, we present the datasets and in Section~\ref{sec:dq}, we analyze their quality based on a method, Data Quality Index, which was previously proposed. Section~\ref{sec:expt}, we give the details of the experimental methods. Section~\ref{sec:results} contains the results and discussion. We conclude in Section~\ref{sec:concl}. 

\section{Related Work on Deception Detection}
Related work can be organized into machine learning approaches, deep learning approaches, and transfer learning approaches to deception detection. 
\subsection{Machine Learning Approaches to Deception Detection}
Text is commonly mapped to some form of numerical data to be used as input (features) for ML models to perform classification upon, and there exist many approaches to extract this data from text. An early work on detecting opinion spam in online reviews utilized several hand-crafted features such as the cosine similarity between the text in a review and a product’s characteristics (as listed in the online description of the product) as well as frequencies of brand names, numerals, and capitals to achieve a 98.7\% AUC value on their data using a logistic regression classifier \cite{jindal2008opinion}. Shojaee et al. \cite{shojaee2013detecting} collect a total of 234 features from text that are a combination of lexical and syntactic elements and use these to analyze a collection of truthful and deceptive online hotel reviews. They implement a Support Vector Machine (SVM) with a Sequential Minimal Optimization (SMO) model as well as a Naïve Bayes (NB) model to achieve an F1 score of 0.74 and 0.84 respectively. Feng et al. \cite{feng2012syntactic} leverage part-of-speech (POS) tags on a collection of truthful and deceptive online reviews and essays to develop parse trees for each text. The parse trees are then combined with unigram frequencies to achieve an average of 74.2\% accuracy across all their datasets using an SVM classifier with 80\% of the data being used for training and 20\% used for testing (80/20 split) and 5-fold cross-validation. Addawood et al. \cite{addawood2019linguistic} propose a variety of potential markers of deception that are tailored towards the environment of Twitter to identify and understand deceptive “troll” behavior around the 2013 U.S. election. They defined trolls as user accounts whose sole purpose is to sow conflict and deception and through statistical analysis determined trolls would use significantly more persuasive language cues as well as a less complex and specific language. They achieved an average F1 score of 0.82 based on the extracted language cues alone. Studies certainly exist that leverage non-text-based features for deception detection. For example, Banerjee et al.\cite{banerjee2014keystroke} utilized keystroke information in combination with term frequency-inverse document frequency (TFIDF) values of unigrams and achieved an accuracy of 94.3\% on a dataset of deceptive and truthful essays using an SVM classifier. Additionally, Crockett et al. \cite{crockett2020automated} use non-verbal head and facial features to identify deceptive or truthful behavior with an accuracy of 99\% using a Random Forest (RF) ensemble classifier. 

\subsection{Deep Learning Approaches to Deception Detection}
The rise of deep learning frameworks has significantly impacted the realm of many NLP tasks as neural network-based models have performed very well in classification tasks. In recent years, deep learning-based models such as convolutional neural networks (CNNs), recurrent neural networks (RNNs), and long short-term memory (LSTM) models have been explored by researchers in various domains of deception detection. Zhao et al. \cite{zhao2018towards} develop a word-order preserving CNN (OPCNN) and evaluate this on a dataset they constructed and annotated. The dataset consisted of over 24,000 online opinions of hotels with approximately 4,100 of those being deceptive. Each text was converted to an embedding using word2vec \cite{mikolov2013efficient} \cite{mikolov2013distributed}, which were used as inputs to the CNN model. They compare the performance of their OPCNN with traditional TFIDF features as well as bigram TFIDF values as classified by an SVM. The OPCNN achieved an accuracy of 70.02\% which is nearly six points higher than that achieved with traditional TFIDF values (64.53\%) and nearly four points higher than the bigram TFIDF model (66.27\%). Their model also outperformed a vanilla CNN which achieved an accuracy of 67.33\% on the data. A disadvantage of a vanilla CNN is that it does not necessarily retain the order of words in a text and therefore can lose the contextual information that previous words may provide. Zhang et al. \cite{zhang2018dri} alleviate this issue by developing a hybrid Recurrent Convolutional Neural Network (RCNN). They employ the Skip-gram \cite{mikolov2013distributed} model to create word vectors that are used as input for the RCNN, and their results show that the inherent temporal component of an RNN can capture the context information of surrounding words and their word embeddings. They evaluate this model on a dataset of deceptive and truthful reviews about hotels, restaurants, and doctors and achieve an accuracy and F1 score of 86\% and 0.84 respectively. A novel method for building upon the Skip-gram model for generating word vectors is presented in \cite{zhang2019dcword}. The authors employ separate Skip-gram models to generate word embeddings separately for truthful and deceptive texts. The final embedding of a word for a given text is the concatenation of the vectors generated by the separate Skip-gram models. An entire text is vectorized by employing an averaging strategy or a max-pooling strategy for all words in the text. Both strategies are evaluated on the same dataset as {19} through several models. They achieve their best accuracy of approximately 90\% using vectors of length 300 combined with the max-pooling strategy, a logistic regression classifier, and using 60\% of the data for training against a randomly sampled 10\% for testing. The remaining 30\% of the data is not used during the experiment. 
Transformer models \cite{vaswani2017attention} have become extremely prevalent in NLP literature and many authors have evaluated them on the task of deception detection. Pre-trained models such as BERT (Bidirectional Encoder Representations from Transformers) \cite{devlin2018bert} have been employed in several papers for the task of deception detection \cite{fornaciari2021bertective} \cite{barsever2020building} \cite{tida2022universal} \cite{gupta2021leveraging} \cite{shahriar2022deception} and have shown to outperform existing state-of-the-art (SOTA) models. However, some research shows that, although successful at capturing semantic information, BERT alone cannot capture the implicit knowledge of deception cues \cite{fornaciari2021bertective}. 

\subsection{Transfer Learning Approaches to Deception Detection}
Commonly, traditional ML relies on large amounts of training data to be available and assumes that the training and testing data are drawn from the same distribution. This assumption does not always hold for many real-world problems, but recent advancements in the field of Transfer Learning (TL) provide strategies to deal with this issue. The primary goal of TL is to solve the target task by leveraging the available data from a source task in a different domain (or domains) \cite{niu2020decade}. TL has been applied to the task of deception detection in a few instances \cite{gupta2021leveraging} \cite{shahriar2022deception} and others have attempted similar work by attempting to develop domain-independent classifiers by evaluating performance on a variety of deceptive domains (i.e., news, Twitter, and phishing) \cite{panda2022deception} \cite{hanks2022data}. These approaches involve identifying domain-invariant features that allow models to better generalize across all deceptive domains \cite{shahriar-etal-2023-exploring} \cite{ng-augmenting} \cite{bazmi-multi-domain}. During our review of existing literature, we found few examples of TL methods being applied to the task of deception detection, and we seek to address this dearth by exploring the effects of a formal TL method used with a variety of classification models and deceptive datasets. 

\section{Datasets} \label{sec:data}
The primary dataset for this study is that created in \cite{banerjee2014keystroke} and will be referred to as the Stony Brook University (SBU) deception dataset. The dataset was developed through crowdsourcing via Amazon Mechanical Turk and is publicly available. Each Turker was directed to a website in which keylogging was enabled and asked to write truthful and deceptive texts with a minimum of 100 words on one of three topics: restaurant review, gay marriage, and gun control. During this time the timings between each key-down event (pressing of a key) and each key-up event (the release of a key) were recorded along with the code for the key itself. Each Turker then wrote two texts for the same topic with one being truthful and one deceptive. For example, the Turker’s opinion was asked for the topics ‘gay marriage’ and ‘gun control’ with the options being support, neutral, or against. Then, they would write one essay that truthfully articulates their opinion on the topic and a second that deceptively presents the opposite opinion. For the ‘restaurant review’ topic, the Turker was asked to write a truthful review of a restaurant they like and have been to, and then write a deceptive review of a restaurant they do not like or have never visited. The dataset is balanced across classes (each Turker writes one truthful and one deceptive) with 1,300 truthful and 1,300 deceptive texts. We consider the dataset as the combination of all topics, and descriptive statistics can be found in Table ~\ref{tab:stats}. One goal of TL is to leverage available data that is in a different, but similar, domain compared to the target dataset. These different domain datasets are referred to as source datasets and are combined with a classification model to impact performance on the target dataset. 

In this study, we consider datasets from several deceptive-adjacent domains: Twitter rumors, fake news, spam text messages, online job scams, phishing emails, and online product reviews. To represent deception from the Twitter domain, we use the PHEME dataset, a corpus of rumor and non-rumor tweets related to various news stories \cite{zubiaga2016learning}. A rumor was a tweet that included information that was unverified at the time of posting. It is not balanced and contains 2,402 rumor tweets and 4,023 non-rumor tweets. A rumor is considered deceptive, and a non-rumor is considered truthful. To represent deception from the news domain, we use the LIAR dataset which is a corpus of short statements from Politifact3 that have been manually labeled \cite{wang2017liar}. The labels provide a range of truth as follows: True, Mostly-True, Half-True, Mostly-False, False, Pants-on-Fire False. We group statements labeled as True and Mostly-True as truthful and the remaining statements as deceptive following similar work \cite{upadhayay2020sentimental}. This results in 6,620 deceptive texts and 3,649 truthful texts. The final deceptive domain we use is the SMS domain which is represented by the SMS Spam Collection dataset. This dataset consists of 4,824 ham (truthful) SMS text messages and 747 spam (deceptive) text messages \cite{almeida2011contributions}. The additional datasets are as follows and have been cleaned and published online for further use: Product Reviews, Job Scams, Phishing, Political Statements, and Fake News \cite{10.1145/3508398.3519358}. Descriptive statistics for all source datasets can be found in Table~\ref{tab:stats} as well. 

\begin{table}
\centering
  \caption{Descriptive Statistics of all Deception Datasets. Length per instance is in terms of number of characters. PS - Political Statements Dataset and PR - Product Reviews Dataset. T- Truthful and D - Deceptive}
  \label{tab:stats}
    \begin{tabular}{|l|l|l|l|}
    \hline
    Dataset& Word Count & Length & $N_{T}$/$N_{D}$\\
    \hline
    PS & 17.7 (7.75)      & 105.12 (44.97) & 7158/5654\\
    Fake News            & 548.13 (3593.29) & 3303.59 (606.18)& 34615/27486\\
    Job Scams            & 186.56 (130.01)  & 1242.37 (873.96)& 13735/608\\
    PR     & 68.8 (84.9)      & 366.79 (468.04) &10481/10481 \\
    Phishing             & 357.65 (857.88)  & 2285.09 (9980.37) & 9202/6134\\
    PHEME                & 15.33 (4.93)     & 99.34 (26.80)& 3830/1972\\
    LIAR                 & 17.91 (7.75)     & 106.32 (45.10)&3649/6620\\
    SMS Spam             & 15.08 (10.91)    & 74.78 (55.10)& 3830/1719\\
    SBU         & 130.18 (40.77)   & 725.11 (241.74)&1300/1300 \\
  \hline
\end{tabular}
\end{table}

\begin{figure}[h]
  \centering
  \includegraphics[width=\linewidth]{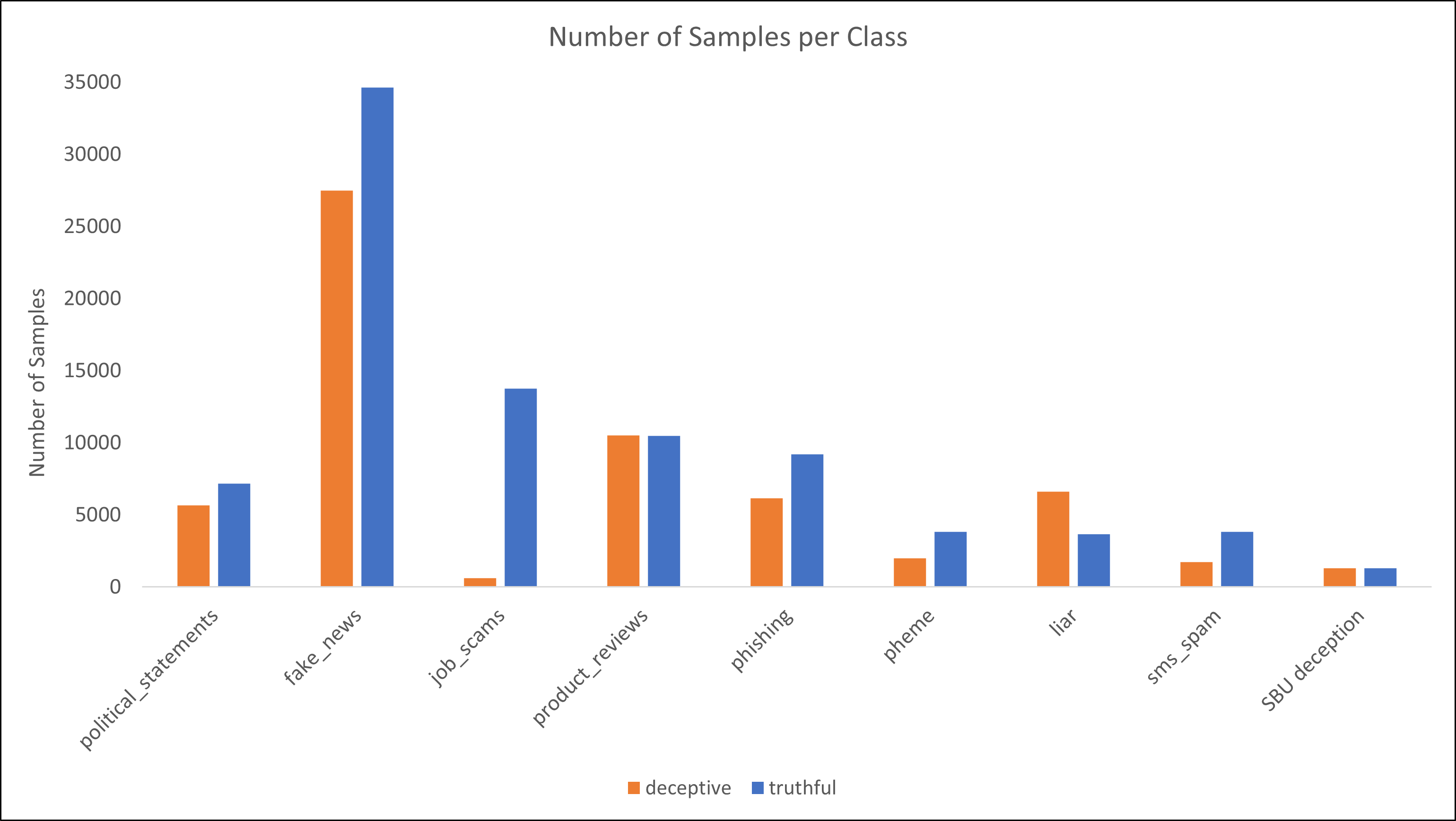}
  \caption{Distribution of Classes for all Deception Datasets}
  \label{fig:stats}
\end{figure}

Each value reported in Table ~\ref{tab:stats} is an average for the dataset and the value in parenthesis is the standard deviation. Additionally, Figure ~\ref{fig:stats} shows the count of each class for all datasets. The SBU deception, Political Statements, and Product Reviews are balanced or only slightly unbalanced. All other datasets are very unbalanced. 

\section{Data Quality Evaluation} \label{sec:dq}
Mishra et al. \cite{mishra2020dqi} propose a generic Data Quality Index (DQI) that can be used to evaluate the quality of datasets used in NLP tasks. The overall DQI is decomposed into seven components which are then further decomposed into multiple sub-terms from which information can be gleaned. Each term is constructed based upon properties of text that are discussed heavily in \cite{mishra2020dqi} and are as follows: 

\begin{enumerate}
	\item Vocabulary  
	\item Inter-sample N-gram Frequency and Relation
	\item Inter-sample Semantic Textual Similarity (STS)
 \item Intra-sample Word Similarity
 \item Intra-sample STS
 \item N-gram Frequency per Label
 \item Inter-spilt STS
\end{enumerate}

The authors apply these to the Stanford Natural Language Inference (SNLI) dataset \cite{bowman-etal-2015-large} to evaluate the DQI and the SNLI dataset is split into a ``good split'' and a ``bad split'' that is determined from applying the adversarial filtering lite (AFLite) \cite{sakaguchi2021winogrande}. These results are included alongside those for the deception datasets to serve as a frame of reference. We did not calculate component five (Intra-sample STS) for the deception datasets as this requires the text of the dataset to have a relationship such as a premise/hypothesis (e.g., the SNLI dataset) or question/answer which all deception datasets in this study do not have. Additionally, two sub-terms of component six were not calculated due to the need for this relationship as well. Due to the scope of this paper, a subset of the complete analysis is presented in this section. All figures not reported in this section will be reported in a longer version of this paper online if this paper is accepted. Additionally, two summary tables of  DQI components are presented in tables ~\ref{tab:DQI_summary123} and tables ~\ref{tab:DQI_summary467}. Due to the difference in dataset labels, the SNLI dataset cannot be directly compared for $DQI_{C6}$. Values that are underlined exceed that of the SNLI Bad Split for the respective DQI component and are considered good quality. Of note, all datasets exceed the SNLI Bad Split for at least one component with most datasets exceeding the SNLI Bad Split for all DQI components. Each sub-term is normalized between zero and one due to the significant difference in scale between sub-terms for $DQI_{C2}$, $DQI_{C3}$, and $DQI_{C6}$.

\begin{table}
\centering
    \caption{Summary of DQI components 1,2 and 3 for all datasets.}
    \label{tab:DQI_summary123}
    \begin{tabular}{|l|l|l|l|}
    \hline
    Dataset               & DQI C1 & DQI C2 & DQI C3 \\
    \hline
    SNLI Good Split       & 7.66   & 0.328  & 0.00314 \\
    SNLI Bad Split        & 6.16   & 0.040  & 0.00583 \\
    SBU Dataset           & \underline{12.68} & \underline{1.762} & \underline{0.07659}\\
    Political Statements & 4.37        & \underline{2.463} & \underline{0.06508} \\
    Fake News            & \underline{17.29} & \underline{5.707} & \underline{1.25669}   \\
    Product Reviews      & \underline{9.34}  & \underline{2.711} & \underline{1.07084} \\
    Job Scams             & \underline{21.69} & \underline{1.200} & \underline{0.63210}  \\
    Phishing              & \underline{39.70} & \underline{3.606} & \underline{0.98167}  \\
    PHEME                 & 2.60        & \underline{2.308} & \underline{0.03748} \\
    LIAR                  & 4.75        & \underline{2.095} & \underline{0.24584} \\
    SMS Spam             & 6.15        & \underline{1.707} & \underline{0.29245}\\
    \hline
\end{tabular}
\end{table}

\begin{table}
\centering
    \caption{Summary of all DQI components 4,6 and 7 for all datasets. $DQI_{C6}$ is marked as N/A for both splits of the SNLI dataset due to the three labels (entailment, neutral, contradiction) not being comparable to the binary deception labels of all other datasets.}
    \label{tab:DQI_summary467}
    \begin{tabular}{|l|l|l|l|}
    \hline
    Dataset               & DQI C4   & DQI C6 & DQI C7 \\
    \hline
    SNLI Good Split        & 0.000372 & N/A    & 0.0031 \\
    SNLI Bad Split         & 0.000062 & N/A    & 0.0029 \\
    SBU Dataset           & \underline{0.017285} & 7.46   & \underline{4.3552} \\
    Political Statements & \underline{0.016652} & 12.09  & \underline{2.3497} \\
    Fake News            & \underline{0.032083} & 9.31   & \underline{7.0734} \\
    Product Reviews      & \underline{0.047775} & 3.81   & \underline{5.1597} \\
    Job Scams            & \underline{0.025044} & 4.84   & \underline{4.0931} \\
    Phishing              & \underline{0.004150} & 1.50   & \underline{5.3857} \\
    PHEME                 & \underline{0.032642} & 5.37   & \underline{2.1974} \\
    LIAR                  & \underline{0.015386} & 5.53   & \underline{2.3773} \\
    SMS Spam             & \underline{0.013593} & 6.89   & \underline{2.7332}\\
    \hline
\end{tabular}
\end{table}

\subsection{DQI Component 1}
The first component $DQI_{C1}$ is defined by Equation ~\ref{eq:c1} and evaluates the size of the vocabulary, distribution of sentence lengths, and the contribution of each sentence to the overall vocabulary. 

\begin{equation} \label{eq:c1}
DQI_{C1} = \frac{v(X)}{size(X)} + \sigma(s(X)) \cdot \frac{\sum_{S} sgn((s-a)(b-s))}{size(S)}
\end{equation}

The variables of Equation ~\ref{eq:c1} are defined as follows:
\begin{itemize}
    \item $X$ = dataset
    \item $v$ = vocabulary
    \item $s$ = sentence length in terms of words
    \item $S$ = set of sentences for dataset
    \item $\sigma$ = standard deviation
    \item $sgn$ = sign of enclosed value
    \item $a,b$ = hyperparameters
\end{itemize}

\begin{figure}[h]
    \centering
    \includegraphics[width=0.8\linewidth]{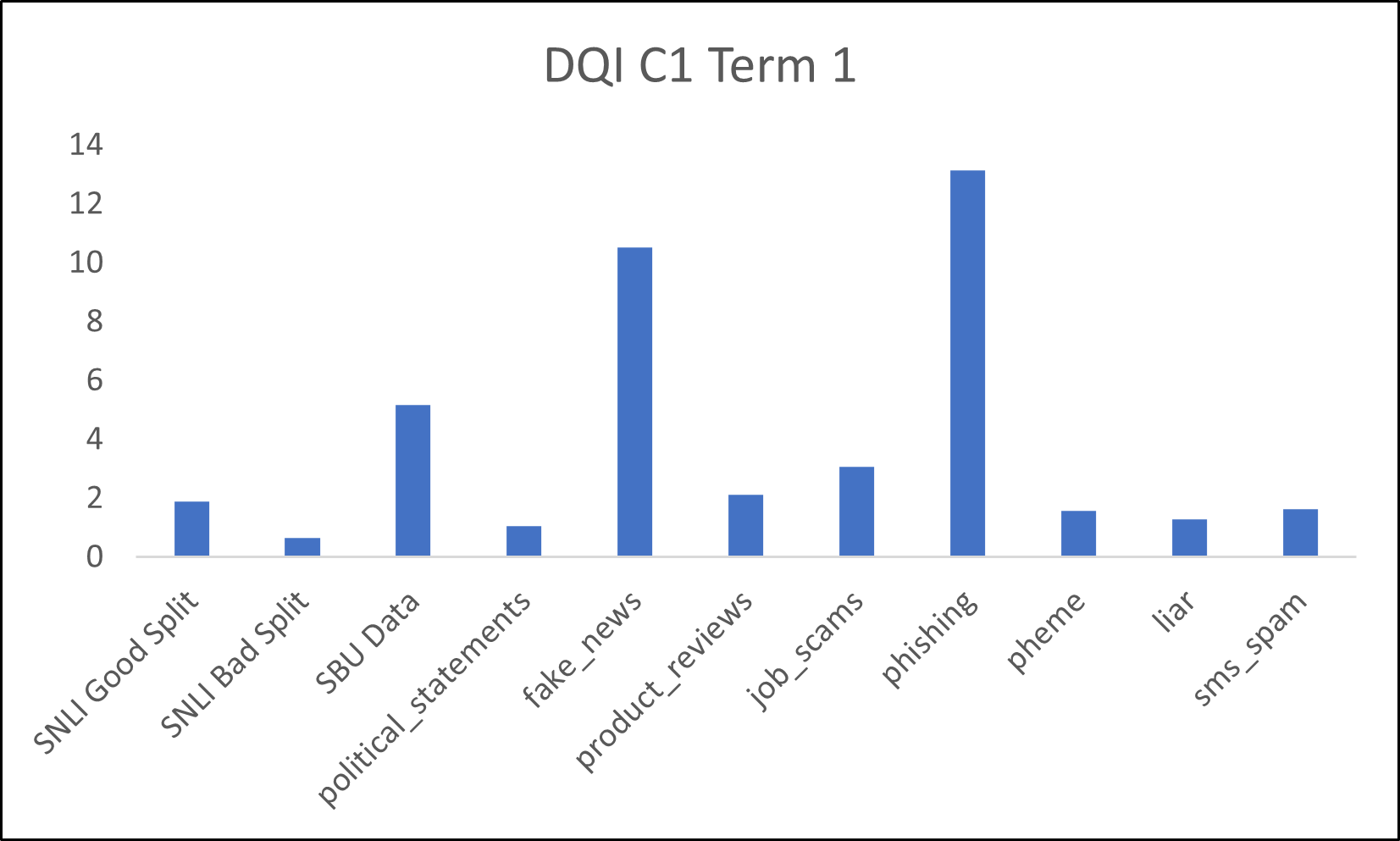}
    \caption{Term 1 of $DQI_{C1}$ for all datasets}
 \label{fig:c1_t1}
\end{figure}

\begin{figure}[h]
    \centering
    \includegraphics[width=0.8\linewidth]{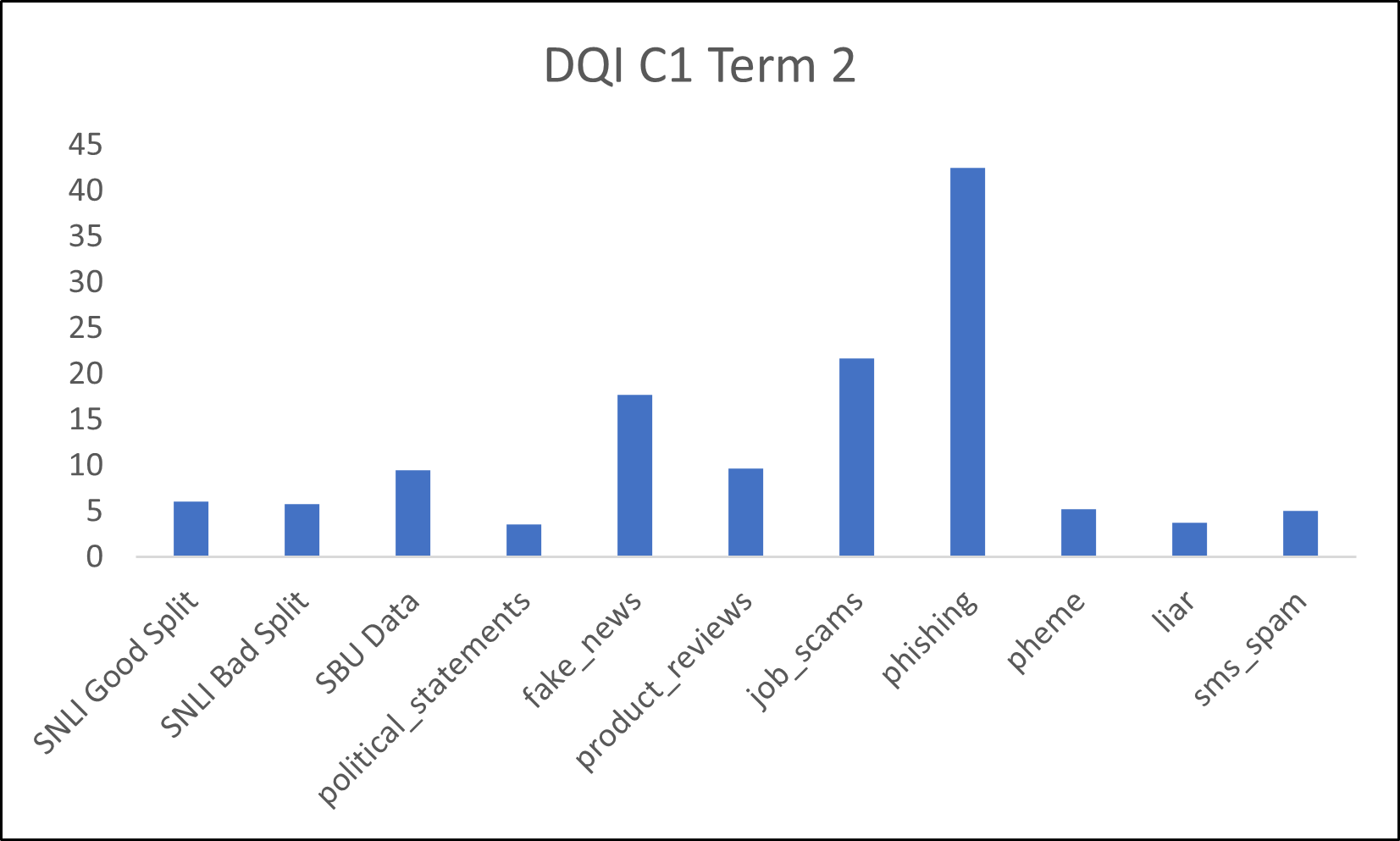}
    \caption{Term 2 of $DQI_{C1}$ for all datasets}
 \label{fig:c1_t2}
\end{figure}

\begin{figure}[h]
    \centering
    \includegraphics[width=0.8\linewidth]{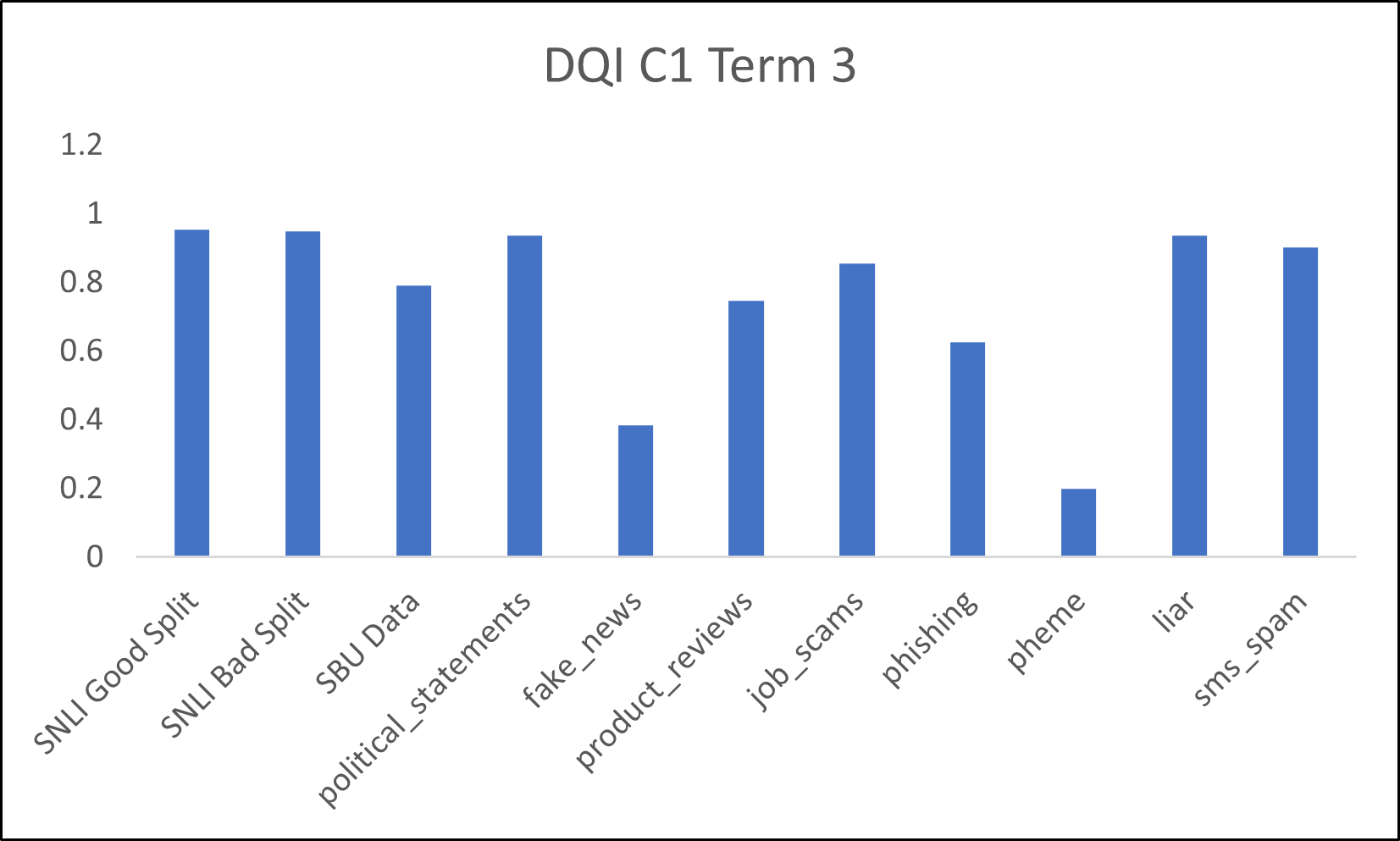}
    \caption{Term 3 of $DQI_{C1}$ for all datasets}
 \label{fig:c1_t3}
\end{figure}

Our analysis used the same values for the hyperparameters a and b as the authors used in \cite{mishra2020dqi}. A larger value is better for all three sub-terms in $DQI_{C1}$ which are shown in Figures ~\ref{fig:c1_t1}, ~\ref{fig:c1_t2}, and ~\ref{fig:c1_t3} respectively. The fake news and phishing datasets notably stand out in sub-term 1 indicating a large vocabulary to dataset size ratio despite the large size of both datasets. Additionally, the SBU dataset value of sub-term 1 is comparably large albeit the relatively small size of the dataset. All other datasets have a term 1 value on a smaller scale. 

For sub-term 2, the phishing dataset has the largest value with all other datasets operating on a similar scale. Of note, sub-term 2 investigates the standard deviation of sentence length in terms of the number of words. We present the standard deviation of sentence length for all datasets in Table ~\ref{tab:stats} in terms of characters, and this is the reason for the difference in values. 

Sub-term 3 investigates the number of sentences that lie near extreme lengths relative to the number of sentences in the dataset. Most datasets operate on a similar scale with the fake news and phishing datasets having notably smaller values than the other datasets. 

\section{Experimental Methods} \label{sec:expt}
Preprocessing consisted of removing any spurious symbols such as newlines and tabs. 

\subsection{Boosting-based Transfer Learning Methods}

The SBU dataset was evaluated using three boosting-based algorithms. Boosting-based algorithms combine multiple weak learners into a strong classifier through an iterative process where each training sample is assigned a weight. After each iteration, the sample weights are updated to assign greater importance to certain samples. AdaBoost \cite{AdaBoost} was used as the baseline, trained only on the target dataset. TrAdaBoost \cite{TrAdaBoost} and gapBoost \cite{gapBoost} were trained on both the source and target datasets. Logistic regression is used as the base learner for all methods, and the number of boosting iterations is set to 10. We found this combination yielded the best balance between model performance and computational efficiency. 90\% of the target dataset is used for training, and the remaining 10\% is reserved for testing.

\subsection{Deep Learning Transfer Method}

\begin{figure}[h]
  \centering
  \includegraphics[width=\linewidth]{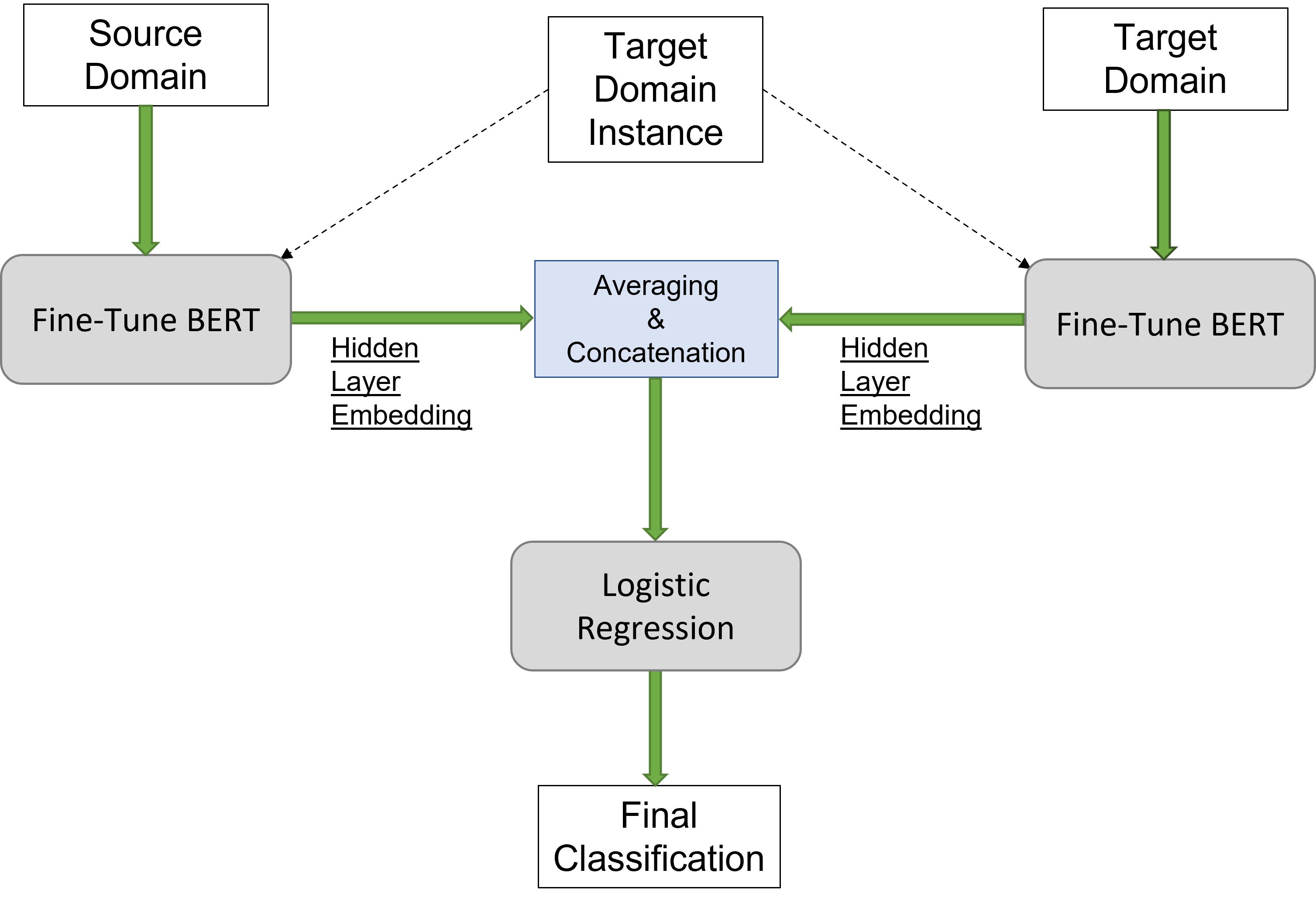}
  \caption{Intermediate Layer Concatenation Process}
  \label{fig:ilc}
\end{figure}

A separate BERT model was fine-tuned for each dataset using the bert-base-uncased implementation found in the HuggingFace transformers library in Python. The following training arguments were used for all fine-tuning: learning rate of 0.00002, weight decay of 0.01, and 5 training epochs. A 90\% sample was used for training with the remaining 10\% used for testing with the class ratio from the dataset being retained in the training and testing sets. Those models trained on datasets that were not the SBU dataset are referred to as source models from this point forward and the model trained on the SBU dataset will be referred to as the target model. After fine-tuning, each source model was used to supplement the target model by performing the intermediate layer concatenation (ILC) as described in \cite{shahriar2022deception} using the hidden layer embeddings. However, a logistic regression (LR) was used as the final classifier instead of a fully connected network due to the model being the top performer of previous baseline experiments. The logistic regression is also trained on a 90\% training sample of the SBU dataset that has been converted into embeddings from the ILC process. A description of the process is found in Figure ~\ref{fig:ilc}. The dashed lines represent the input of a target instance for classification. 

Additionally, variants of all datasets were created in which information was added based on named entities (NEs). This was accomplished using the pipeline available in the Stanza package in Python. The following four methods were used which added or replaced text in the dataset:

\begin{itemize}
\item Method 1: Replace NE with an explanation
\item Method 2: Replace NE with part of speech (POS) tag
\item Method 3: Attach explanation to NE
\item Method 4: Attach POS tag to NE
\end{itemize}

\subsection{Distance Measurements}

Multiple distance measurements were employed to evaluate the similarity between the source datasets and the target dataset. The Kullback-Leibler (KL) divergence ~\cite{kullback1951information} and Jensen-Shannon (JS) distance ~\cite{lin1991divergence} were calculated between each source dataset and the target dataset (SBU dataset). KL divergence is defined below in Equation ~\ref{eq:kl_div} where $P$ and $Q$ represent the source and target dataset distributions respectively, and $p_i$ and $q_i$ are instances of their respective dataset's distribution. KL divergence is not a symmetric measurement, therefore $D_{KL}(P||Q)$ and $D_{KL}(Q||P)$ are both calculated.

\begin{equation} \label{eq:kl_div}
    D_{KL}(P||Q) = \sum_{i=1}^n p_i \log\frac{p_i}{q_i}
\end{equation}

JS distance is defined in Equation ~\ref{eq:js_dis} and employs the KL divergence between the source and target distribution to $M$ which is defined as $M = \frac{1}{2}(P+Q)$, i.e. the average distribution between $P$ and $Q$

\begin{equation} \label{eq:js_dis}
    D_{JS} = \frac{1}{2} [D_{KL}(P||M) + D_{KL}(Q||M)] 
\end{equation}

Probability distributions are calculated from inverse-document frequency (IDF) values for each dataset's vocabulary. IDF values are smoothed by adding one to prevent division by zero. Finally, a cosine-based similarity measurement was evaluated as well which was initially proposed by Panda and Levitan ~\cite{panda2022deception} and is defined in Equation ~\ref{eq:cos_sim} where $D_S$ and $D_T$ are the source and target datasets, and $SD_S$ and $SD_T$ are the average sentence embeddings for all sentences in the source and target datasets respectively. Sentence embeddings were obtained using the sentence-transformer model all-distilroberta-v1. 

\begin{equation} \label{eq:cos_sim}
    D_{cos}(D_S, D_T) = \frac{1-cos(SD_S, SD_T)}{2}
\end{equation}

\section{Results and Discussion} \label{sec:results}
\subsection{Boosting-based Methods}

\begin{figure}[h]
  \centering
  \includegraphics[width=\linewidth]{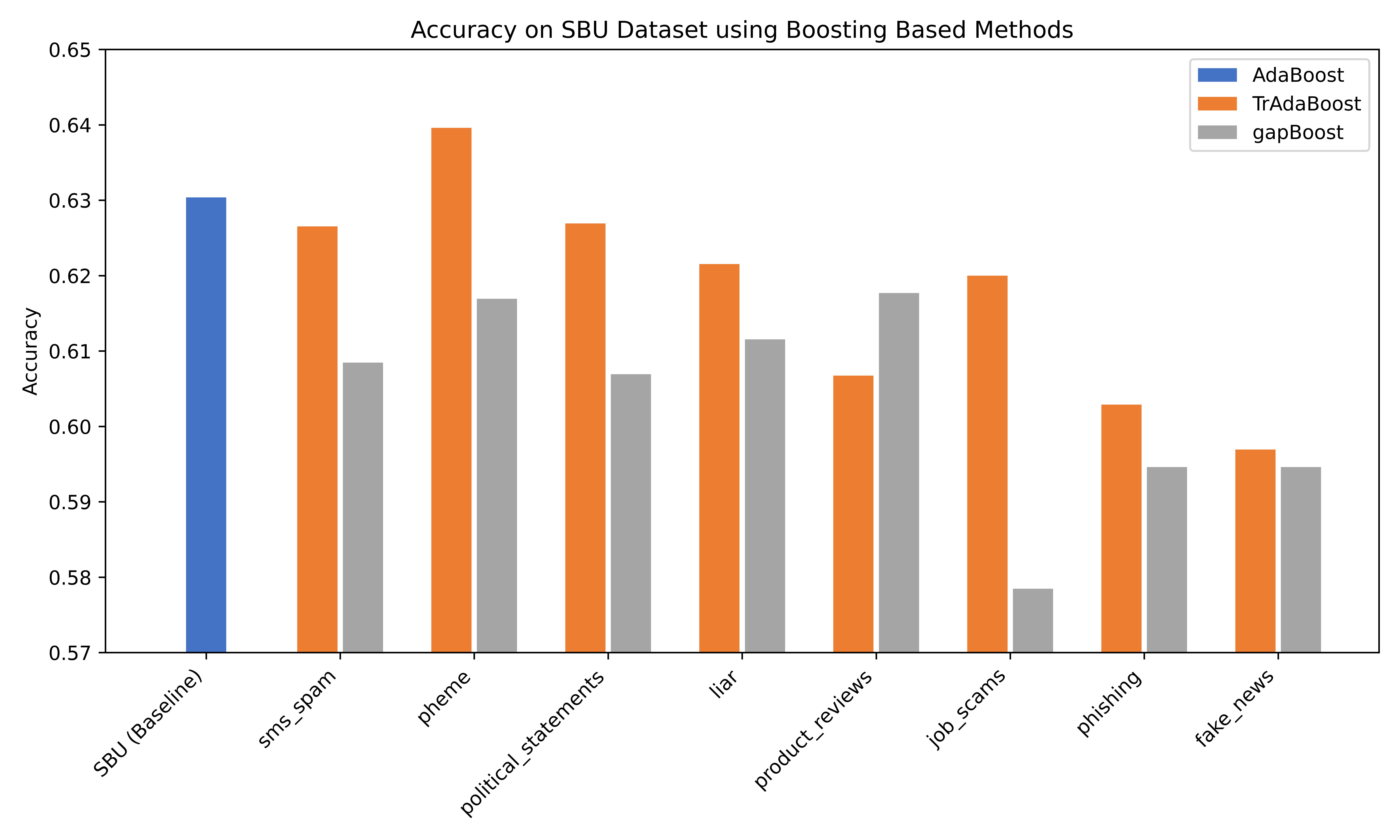}
  \caption{Effect of boosting-based transfer learning on deception detection accuracy of SBU dataset compared to baseline performance}
  \label{fig:boosting}
\end{figure}

The results of the boosting-based transfer learning methods can be seen in Figure ~\ref{fig:boosting} with AdaBoost serving as the baseline accuracy. There is only one instance of improvement over the baseline, with TrAdaBoost improving over the baseline by 1.46\% with PHEME as the source dataset. A possible explanation for these results is that the PHEME datset may have certain characteristics that are more similar to the target SBU dataset compared to the other source datasets. This similarity could allow for more effective knowledge transfer. However, this seems unlikely, given that the PHEME datasets ranks lower in similarity to the target dataset based on the distance metrics presented in Table \ref{tab:distances}. A more plausible explanation is the imbalance in dataset sizes: many of the source datasets are significantly larger than the target SBU dataset, which may have caused the target dataset samples to be overshadowed, leading to negative transfer. This phenomenon is further discussed in Section 5.4.


\subsection{Intermediate Layer Concatenation}
\begin{figure}[h]
  \centering
  \includegraphics[width=\linewidth]{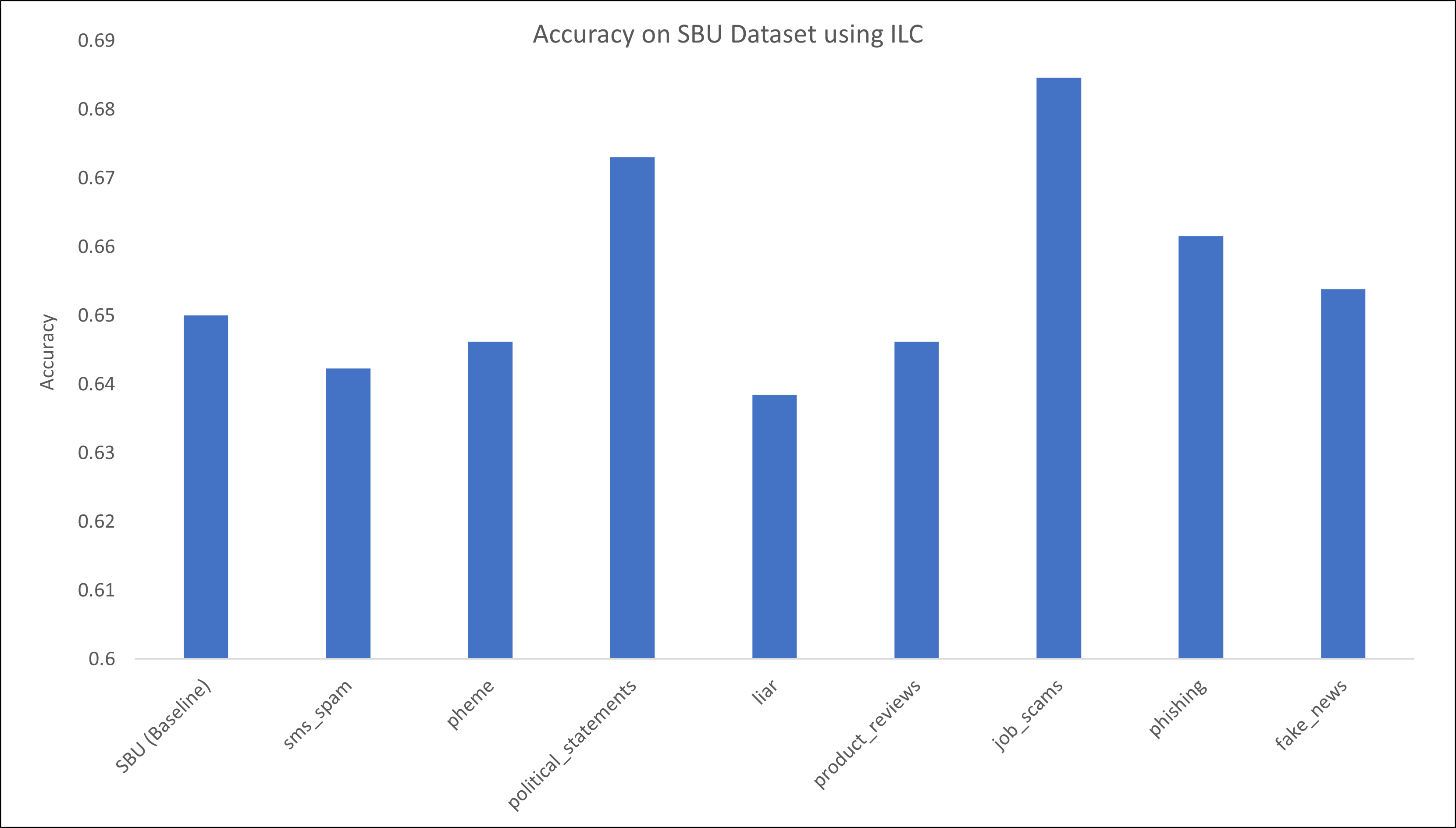}
  \caption{Effect of ILC transfer learning on deception detection accuracy of SBU dataset compared to baseline performance}
  \label{fig:ilc_results}
\end{figure}

The effect of transfer learning via ILC is shown in Figure ~\ref{fig:ilc_results} which includes the accuracy of a BERT model that is fine-tuned on the SBU dataset alone and is referred to as the baseline BERT accuracy. Each source dataset indicates the use of that source model in the ILC process. Four datasets showed improvement over baseline BERT accuracy of 65\%. Fake News showed the smallest improvement of 0.59\% and the Phishing dataset resulted in an improvement of 1.77\%. The Political Statements dataset showed an improvement of 3.55\% and the Job Scams dataset was the most beneficial with an improvement of 5.32\%. 

\begin{figure}[h]
  \centering
  \includegraphics[width=\linewidth]{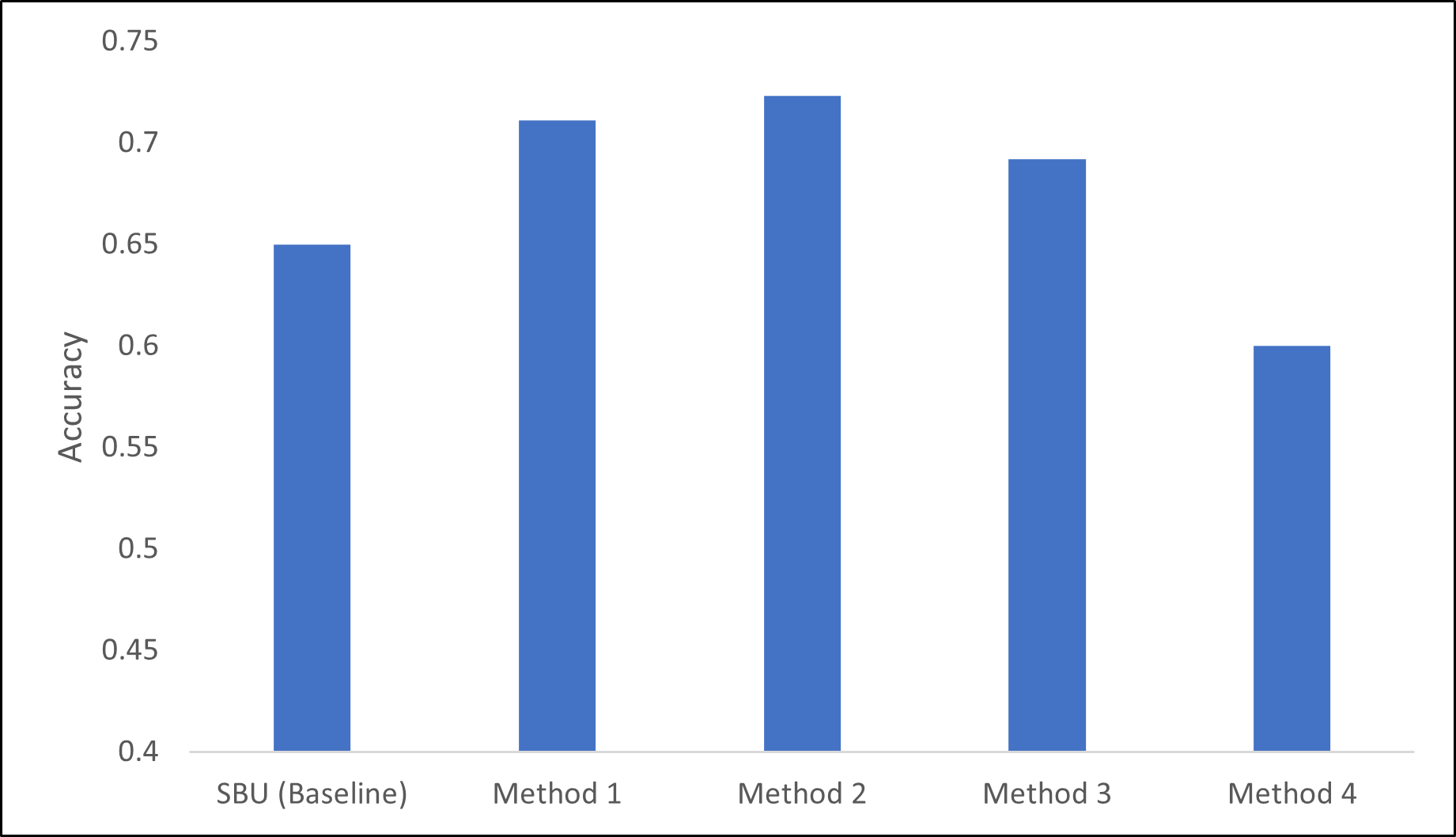}
  \caption{Effect of altering named-entity information on BERT performance compared to baseline}
  \label{fig:sbu_elk}
\end{figure}

The effect of altering the SBU dataset using the four NE methods was explored by fine-tuning a BERT model on each respective NE version of the SBU dataset. This was done to develop a baseline effect of including NE information on a target dataset and is presented in Figure \ref{fig:sbu_elk} with methods one, two, and three resulting in improved accuracy. Method three yielded the smallest improvement of 6.46\%. Method one yielded an improvement of 9.38\% and method two resulted in the largest improvement of 11.2\% over baseline BERT performance. An explanation for this could be that the replacement of NEs with POS tags results in text that is more homogenized since there are likely fewer POS tags than NEs. After replacement, the word frequency distribution should be change noticeably, for example instead of many texts discussing different restaurants this would become the POS tag ``noun'' for each mention of the restaurant. Doing this at the scale of the entire dataset may make separating the deceptive texts easier by forcing the machine learning methods to focus on linguistic cues other than NEs.

\begin{figure}[h]
  \centering
  \includegraphics[width=\linewidth]{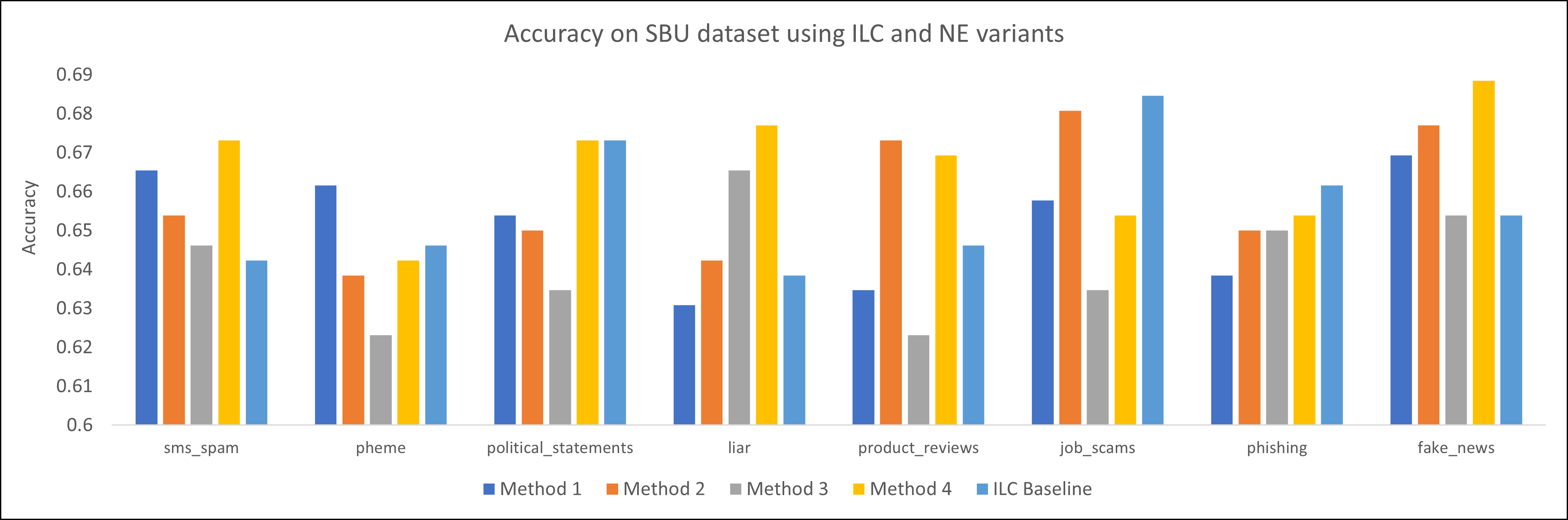}
  \caption{Effect of ILC transfer learning on deception detection accuracy of SBU dataset when named-entity information is altered}
  \label{fig:ilc_elk}
\end{figure}

The ILC experiments were also performed using all variants created using the NE information and are presented in Figure ~\ref{fig:ilc_elk}. For these experiments, a separate BERT model was fine-tuned for each variation of each dataset and only datasets altered via the same method would be compared. For example, a source model trained on the LIAR dataset altered using method one would be paired with a target model trained on the SBU dataset altered using method one as well. The ILC baseline results in Figure ~\ref{fig:ilc_elk} are the same as those found in Figure ~\ref{fig:ilc_results} and are included as a reference for the effect of the NE variants. For example, the ILC baseline value in Figure ~\ref{fig:ilc_elk} for the SMS Spam dataset is the same result for the SMS Spam dataset in Figure ~\ref{fig:ilc_results}. The Political Statements, Job Scams, and Phishing datasets showed no increase in accuracy when compared to ILC baseline for any of the methods. However, all other datasets resulted in an increase in accuracy when compared to their baseline ILC performance with the largest increase being a 5.29\% increase using the LIAR dataset. Furthermore, every dataset showed improvement for at least one NE variant when compared to the baseline BERT accuracy of 65\% achieved by fine-tuning on the SBU dataset alone. 

\subsection{Distance Measurements}

The results for distance measurements are in table ~\ref{tab:distances} and show varying levels of similarity. The Fake News and Phishing datasets are by far the most similar to the SBU dataset as they have the lowest KL divergences and lowest JS distance however they have a greater cosine distance. Other source datasets show similar KL divergence and JS distance values and are generally within 10\% of each other. However, using the cosine-based distance the Political Statements and LIAR datasets are the most similar to the SBU dataset. 

\begin{table}[]
\centering
\caption{Values for distance measures between target dataset $Q$ and all source datasets $P$.}
\label{tab:distances}
\begin{tabular}{|l|l|l|l|l|}
\hline
Source Dataset        & $D_{KL}(Q||P)$ & $D_{KL}(P||Q)$ & $D_{JS}$ & $D_{cos}$ \\
\hline
Political Statements & 1.140           & 1.062           & 0.228   &  0.287\\
Fake News            & 0.703           & 0.392           & 0.114   &  0.307\\
Product Reviews      & 1.096           & 0.809           & 0.199   &  0.356\\
Job Scams            & 1.217           & 0.929           & 0.224   &  0.423\\
Phishing             & 0.689           & 0.353           & 0.105   &  0.356\\
PHEME                & 1.106           & 1.208           & 0.243   &  0.315\\
LIAR                 & 1.132           & 1.082           & 0.230   &  0.281\\
SMS Spam             & 1.226           & 1.327           & 0.267   &  0.446\\
\hline
\end{tabular}
\end{table}

The Pearson correlation coefficient ($r$) was calculated between each set of distance measurements and the change in accuracy that resulted from the ILC experiments performed in Figure ~\ref{fig:ilc_results}. Furthermore, the Spearman rank correlation coefficient ($\rho$) was also calculated between each set of distance measurements and the change in accuracy that resulted from the ILC experiments. These results are in Table ~\ref{tab:correlations} and show that KL Divergence ($D_{KL}(P||Q)$) and JS distance show moderate correlation for both the Pearson and Spearman coefficients. The largest correlation is a Spearman value of 0.527 for both KL Divergence ($D_{KL}(P||Q)$) and JS distance with identical values resulting from all source datasets receiving the same rank regardless of the two distance measurements. Distance measurements were ranked in ascending order for a change in accuracy from ILC and descending order for distance measurement to measure the correlation between the increase in performance and decrease in distance (i.e., increase in similarity). 

\begin{table}[]
\centering
\caption{Correlations between distance measurements and change in accuracy from ILC.}
\label{tab:correlations}
\begin{tabular}{|l|l|l|l|l|}
\hline
Correlation        & $D_{KL}(Q||P)$ & $D_{KL}(P||Q)$ & $D_{JS}$ & $D_{cos}$ \\
\hline
$r$                 &0.214	        &0.369	      & 0.328	    &-0.219\\
$\rho$              & 0.060	        &0.527        &0.527	    &-0.120\\
\hline
\end{tabular}
\end{table}

\subsection{Boosting with Reduced Source Datasets}
Given the relatively small size of the SBU dataset compared to some of the other source datasets seen in Table~\ref{tab:stats}, we hypothesized that an overabundance of source dataset samples could overshadow the target samples and contribute to negative transfer. To investigate this idea, we ran the same boosting experiments on a reduced version of each source dataset composed of 100 randomly selected samples, with 50 from each class.

\begin{figure}[h]
  \centering
  \includegraphics[width=\linewidth]{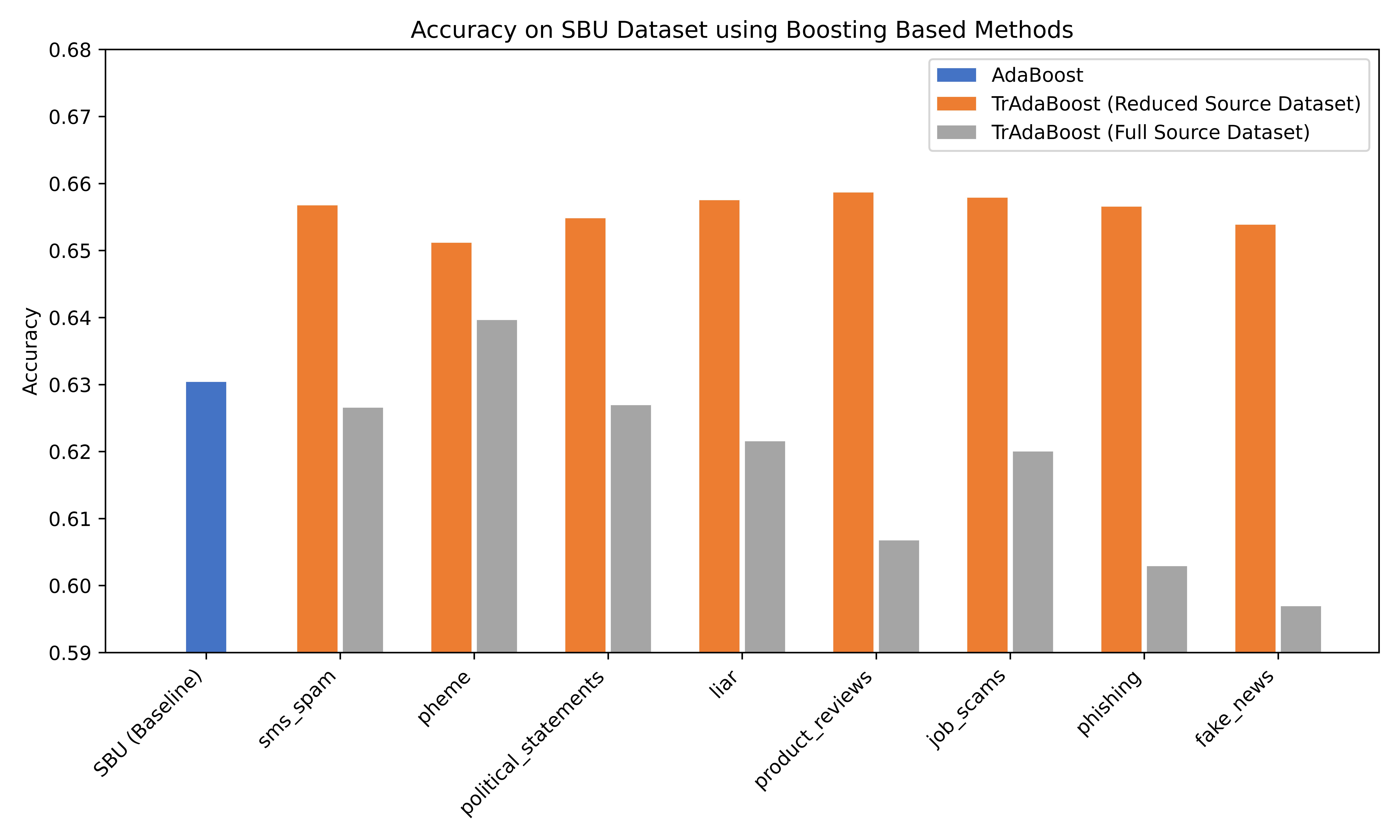}
  \caption{Effect of Reducing Source Dataset Size on TrAdaBoost deception detection accuracy of SBU dataset}
  \label{fig:tr_boosting_reduced}
\end{figure}

\begin{figure}[h]
  \centering
  \includegraphics[width=\linewidth]{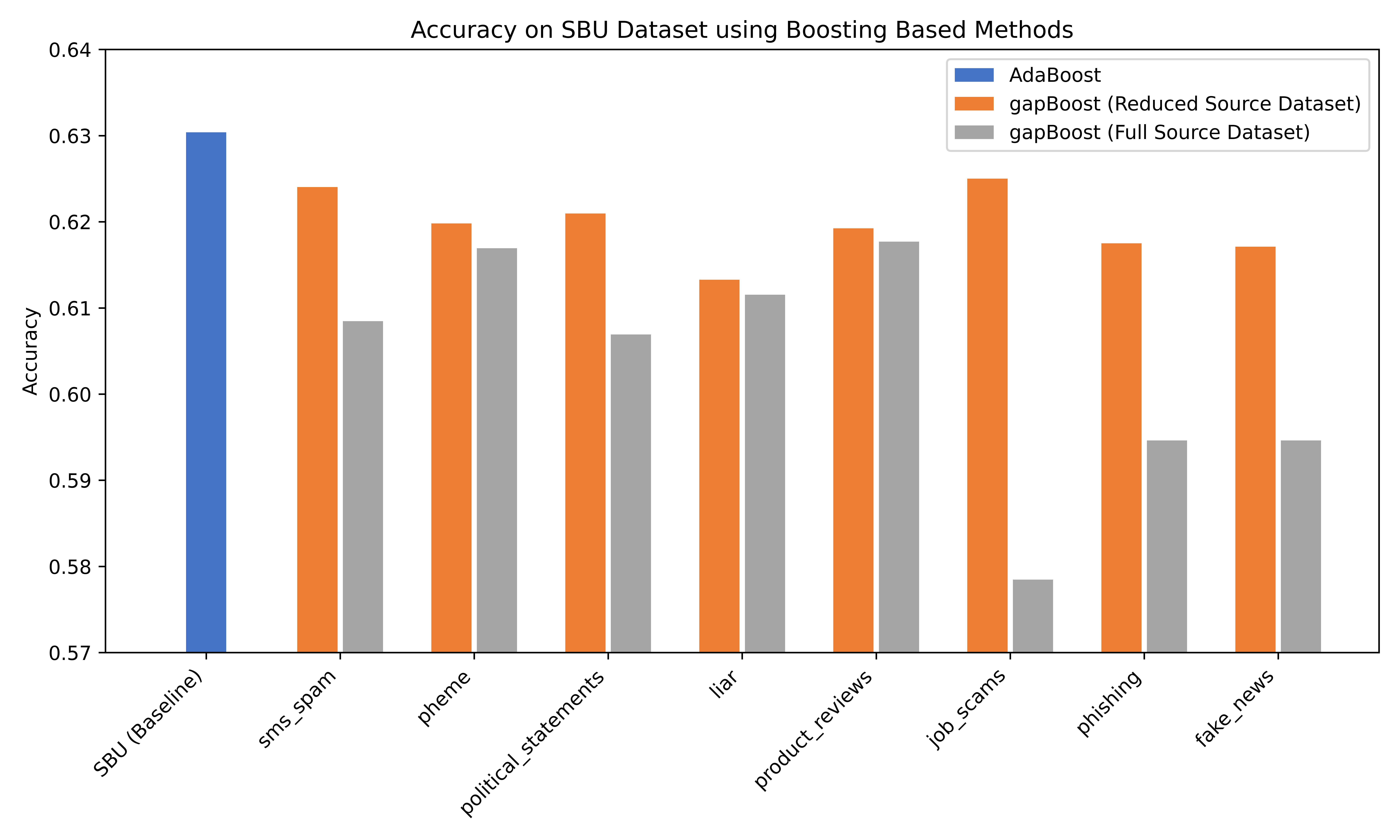}
  \caption{Effect of Reducing Source Dataset Size on gapBoost deception detection accuracy of SBU dataset}
  \label{fig:gb_boosting_reduced}
\end{figure}

Preliminary results shown in Figure ~\ref{fig:tr_boosting_reduced} and ~\ref{fig:gb_boosting_reduced} are promising, with TrAdaBoost exhibiting improvement over the baseline in all eight source datasets. Specifically, the reduced Product Reviews dataset shows the largest improvement over the baseline of 8.56\%. The largest increase over a full source dataset is observed with Fake News, where the reduced dataset improves by 9.54\% over the full dataset.

Although gapBoost still does not outperform the baseline, there are significant increases in accuracy compared to the full source dataset results. The largest such increase is seen with the Job Scams dataset, where an 8.05\% increase in accuracy is observed.

\subsection{Adversarial Considerations}
Adversarial behavior must be considered given this paper focuses on classifying deceptive intent. All BERT based systems developed in this work will suffer from known vulnerabilities of this model, however the focus of this discussion is on the methods of this study as opposed to the specific LLM used for feature extraction and classification, which is practically interchangeable among the plethora of encoder-decoder LLMs making it difficult for an adversary to develop a model-specific attack. In a white-box scenario, an adversary would be aware of all details for the methodology and could change writing patterns to avoid detection. However, as previously discussed, one goal of this work is to investigate the effect of transfer learning which, at a practical level, involved supplementing a target dataset with other deceptive data. One benefit of this system is that the source of supplemented data could be changed depending on the adversaries new style. Additionally, the most promising method (with respect to the SBU dataset) was the replacement of named entities with their POS tag. This would require adversaries to completely avoid using NEs in text which, although not impossible, is a difficult task to do.

\section{Conclusion} \label{sec:concl}

In this work, we performed an extensive data quality analysis for several publicly available deception datasets using the DQI approach and compared the results to the existing analysis performed for the SNLI dataset. We found that, generally, the deception datasets we used in this study meet or exceed the quality measurements of the established values for the SNLI dataset. We evaluated the effect of two boosting-based transfer learning methods for deception detection and found that, except the PHEME source dataset, performance decreased, however, reducing the size of the source dataset improved performance for both methods. Additionally, we evaluated how effective BERT is for classification on the SBU dataset to establish a baseline for deep-learning based deception detection. Furthermore, we showed that the ILC approach can outperform baseline BERT performance for multiple source datasets and boosting-based transfer learning methods with the Job Scams dataset being the most useful. We calculated KL divergence, JS distance, and a cosine-based distance between all source datasets and the SBU dataset to function as similarity measurements and determined that KL divergence and JS distance have a moderate correlation with the ILC transfer learning method. Finally, we determined that NE variants of the target dataset can improve baseline BERT performance and NE variants of source datasets can improve upon ILC performance with most source datasets showing an increase in performance when compared to the ILC baseline. 

For future work, the effects of additional transfer learning methods should be explored as well as additional source datasets. Furthermore, due to limits in time, each BERT model was fine-tuned on only one dataset, however, it is commonly known that deep learning models benefit from larger amounts of training data. Therefore, it would be prudent to explore the effect that ILC has when the source model has been fine-tuned on multiple datasets. Finally, it would be worthwhile to evaluate different network architectures for final classification in ILC.  

\textbf{Acknowledgements} Research partly supported by NSF grants 2210198 and 2244279,  ARO grants W911NF-20-1-0254 and W911NF-23-1-0191, and a USDOT Cyber transportation center grant.
\textbf{Disclosure of Interests}
Verma is the founder of Everest Cyber Security and Analytics, Inc.

%
%
%
\bibliographystyle{splncs04}
\bibliography{bibliography}

\begin{thebibliography}{10}
\providecommand{\url}[1]{\texttt{#1}}
\providecommand{\urlprefix}{URL }
\providecommand{\doi}[1]{https://doi.org/#1}

\bibitem{addawood2019linguistic}
Addawood, A., Badawy, A., Lerman, K., Ferrara, E.: Linguistic cues to deception: Identifying political trolls on social media. In: Proceedings of the international AAAI conference on web and social media. vol.~13, pp. 15--25 (2019)

\bibitem{almeida2011contributions}
Almeida, T.A., Hidalgo, J.M.G., Yamakami, A.: Contributions to the study of sms spam filtering: new collection and results. In: Proceedings of the 11th ACM symposium on Document engineering. pp. 259--262 (2011)

\bibitem{banerjee2014keystroke}
Banerjee, R., Feng, S., Kang, J.S., Choi, Y.: Keystroke patterns as prosody in digital writings: A case study with deceptive reviews and essays. In: Proceedings of the 2014 conference on empirical methods in natural language processing (EMNLP). pp. 1469--1473 (2014)

\bibitem{barsever2020building}
Barsever, D., Singh, S., Neftci, E.: Building a better lie detector with {BERT}: The difference between truth and lies. In: 2020 International Joint Conference on Neural Networks (IJCNN). pp.~1--7. IEEE (2020)

\bibitem{bazmi-multi-domain}
Bazmi, P., Asadpour, M., Shakery, A., Maazallahi, A.: Entity-centric multi-domain transformer for improving generalization in fake news detection. Information Processing \& Management  \textbf{61}(5),  103807 (2024). \doi{https://doi.org/10.1016/j.ipm.2024.103807}, \url{https://www.sciencedirect.com/science/article/pii/S0306457324001663}

\bibitem{bowman-etal-2015-large}
Bowman, S.R., Angeli, G., Potts, C., Manning, C.D.: A large annotated corpus for learning natural language inference. In: Proceedings of the 2015 Conference on Empirical Methods in Natural Language Processing. pp. 632--642. Association for Computational Linguistics, Lisbon, Portugal (Sep 2015). \doi{10.18653/v1/D15-1075}, \url{https://aclanthology.org/D15-1075}

\bibitem{burgoon1994interpersonal}
Burgoon, J.K., Buller, D.B.: Interpersonal deception: Iii. effects of deceit on perceived communication and nonverbal behavior dynamics. Journal of Nonverbal Behavior  \textbf{18},  155--184 (1994)

\bibitem{crockett2020automated}
Crockett, K., O’Shea, J., Khan, W.: Automated deception detection of males and females from non-verbal facial micro-gestures. In: 2020 International Joint Conference on Neural Networks (IJCNN). pp.~1--7. IEEE (2020)

\bibitem{TrAdaBoost}
Dai, W., Yang, Q., Xue, G.R., Yu, Y.: Boosting for transfer learning. In: Proc. 24th ICML. p. 193–200 (2007)

\bibitem{devlin2018bert}
Devlin, J., Chang, M.W., Lee, K., Toutanova, K.: Bert: Pre-training of deep bidirectional transformers for language understanding. arXiv preprint arXiv:1810.04805  (2018)

\bibitem{egozi2018phishing}
Egozi, G., Verma, R.: Phishing email detection using robust nlp techniques. In: 2018 IEEE International Conference on Data Mining Workshops (ICDMW). pp. 7--12. IEEE (2018)

\bibitem{aasalBD2020}
El~Aassal, A., Baki, S., Das, A., Verma, R.M.: An in-depth benchmarking and evaluation of phishing detection research for security needs. IEEE Access  \textbf{8},  22170--22192 (2020). \doi{10.1109/ACCESS.2020.2969780}

\bibitem{feng2012syntactic}
Feng, S., Banerjee, R., Choi, Y.: Syntactic stylometry for deception detection. In: Proceedings of the 50th Annual Meeting of the Association for Computational Linguistics (Volume 2: Short Papers). pp. 171--175 (2012)

\bibitem{fornaciari2021bertective}
Fornaciari, T., Bianchi, F., Poesio, M., Hovy, D., et~al.: Bertective: Language models and contextual information for deception detection. In: Proceedings of the 16th Conference of the European Chapter of the Association for Computational Linguistics: Main Volume. Association for Computational Linguistics (2021)

\bibitem{AdaBoost}
Freund, Y., Schapire, R.E.: A desicion-theoretic generalization of on-line learning and an application to boosting. In: Vit{\'a}nyi, P. (ed.) Computational Learning Theory. pp. 23--37. Springer Berlin Heidelberg, Berlin, Heidelberg (1995)

\bibitem{gupta2021leveraging}
Gupta, P., Gandhi, S., Chakravarthi, B.R.: Leveraging transfer learning techniques-{BERT}, {RoBERTta}, {ALBERT} and {DistilBERT} for fake review detection. In: Proc. of the 13th FIRE meeting. pp. 75--82 (2021)

\bibitem{hanks2022data}
Hanks, C., Verma, R.M.: Data quality and linguistic cues for domain-independent deception detection. In: 2022 IEEE/ACM International Conference on Big Data Computing, Applications and Technologies (BDCAT). pp. 248--258. IEEE (2022)

\bibitem{hauch2015computers}
Hauch, V., Bland{\'o}n-Gitlin, I., Masip, J., Sporer, S.L.: Are computers effective lie detectors? a meta-analysis of linguistic cues to deception. Personality and social psychology Review  \textbf{19}(4),  307--342 (2015)

\bibitem{jindal2008opinion}
Jindal, N., Liu, B.: Opinion spam and analysis. In: Proceedings of the 2008 international conference on web search and data mining. pp. 219--230 (2008)

\bibitem{kullback1951information}
Kullback, S., Leibler, R.A.: On information and sufficiency. The annals of mathematical statistics  \textbf{22}(1),  79--86 (1951)

\bibitem{lazer2018science}
Lazer, D.M., Baum, M.A., Benkler, Y., Berinsky, A.J., Greenhill, K.M., Menczer, F., Metzger, M.J., Nyhan, B., Pennycook, G., Rothschild, D., et~al.: The science of fake news. Science  \textbf{359}(6380),  1094--1096 (2018)

\bibitem{li2021exploring}
Li, J., Lv, P., Xiao, W., Yang, L., Zhang, P.: Exploring groups of opinion spam using sentiment analysis guided by nominated topics. Expert Systems with Applications  \textbf{171},  114585 (2021)

\bibitem{lin1991divergence}
Lin, J.: Divergence measures based on the shannon entropy. IEEE Transactions on Information theory  \textbf{37}(1),  145--151 (1991)

\bibitem{liu2020detecting}
Liu, M., Shang, Y., Yue, Q., Zhou, J.: Detecting fake reviews using multidimensional representations with fine-grained aspects plan. IEEE Access  \textbf{9},  3765--3773 (2020)

\bibitem{mikolov2013efficient}
Mikolov, T., Chen, K., Corrado, G., Dean, J.: Efficient estimation of word representations in vector space. arXiv preprint arXiv:1301.3781  (2013)

\bibitem{mikolov2013distributed}
Mikolov, T., Sutskever, I., Chen, K., Corrado, G.S., Dean, J.: Distributed representations of words and phrases and their compositionality. NIPS  \textbf{26} (2013)

\bibitem{mishra2020dqi}
Mishra, S., Arunkumar, A., Sachdeva, B., Bryan, C., Baral, C.: Dqi: Measuring data quality in nlp. arXiv preprint arXiv:2005.00816  (2020)

\bibitem{ng-augmenting}
Ng, K.C., Ke, P.F., So, M.K.P., Tam, K.Y.: Augmenting fake content detection in online platforms: A domain adaptive transfer learning via adversarial training approach. Production and Operations Management  \textbf{32}(7),  2101--2122 (2023). \doi{10.1111/poms.13959}, \url{https://doi.org/10.1111/poms.13959}

\bibitem{niu2020decade}
Niu, S., Liu, Y., Wang, J., Song, H.: A decade survey of transfer learning (2010--2020). IEEE Transactions on Artificial Intelligence  \textbf{1}(2),  151--166 (2020)

\bibitem{panda2022deception}
Panda, S., Levitan, S.: Deception detection within and across domains: Identifying and understanding the performance gap. ACM Journal of Data and Information Quality  \textbf{15}(1),  1--27 (2022)

\bibitem{sakaguchi2021winogrande}
Sakaguchi, K., Bras, R.L., Bhagavatula, C., Choi, Y.: Winogrande: An adversarial winograd schema challenge at scale. Communications of the ACM  \textbf{64}(9),  99--106 (2021)

\bibitem{shahriar2022deception}
Shahriar, S., Mukherjee, A., Gnawali, O.: Deception detection with feature-augmentation by soft domain transfer. In: Int'l Conf. on Social Informatics. pp. 373--380 (2022)

\bibitem{shahriar-etal-2023-exploring}
Shahriar, S., Mukherjee, A., Gnawali, O.: Exploring deceptive domain transfer strategies: Mitigating the differences among deceptive domains. In: Mitkov, R., Angelova, G. (eds.) Proc. 14th RANLP. pp. 1076--1084 (Sep 2023)

\bibitem{shojaee2013detecting}
Shojaee, S., Murad, M.A.A., Azman, A.B., Sharef, N.M., Nadali, S.: Detecting deceptive reviews using lexical and syntactic features. In: 2013 13th International Conference on Intellient Systems Design and Applications. pp. 53--58. IEEE (2013)

\bibitem{tang2020review}
Tang, H., Cao, H.: A review of research on detection of fake commodity reviews. In: Journal of Physics: Conference Series. vol.~1651, p. 012055 (2020)

\bibitem{tida2022universal}
Tida, V.S., Hsu, S.: Universal spam detection using transfer learning of {BERT} model. arXiv preprint arXiv:2202.03480  (2022)

\bibitem{upadhayay2020sentimental}
Upadhayay, B., Behzadan, V.: Sentimental liar: Extended corpus and deep learning models for fake claim classification. In: 2020 IEEE ISI Conf.). pp.~1--6 (2020)

\bibitem{vaswani2017attention}
Vaswani, A., Shazeer, N., Parmar, N., Uszkoreit, J., Jones, L., Gomez, A.N., Kaiser, {\L}., Polosukhin, I.: Attention is all you need. Advances in neural information processing systems  \textbf{30} (2017)

\bibitem{verma2024}
Verma, R.M., Dershowitz, N., Zeng, V., Boumber, D., Liu, X.: Domain-independent deception: A new taxonomy and linguistic analysis (2024), \url{https://arxiv.org/abs/2402.01019}

\bibitem{gapBoost}
Wang, B., Mendez, J., Cai, M., Eaton, E.: Transfer learning via minimizing the performance gap between domains. In: NIPS. vol.~32 (2019)

\bibitem{wang2012identify}
Wang, G., Xie, S., Liu, B., Yu, P.S.: Identify online store review spammers via social review graph. ACM Transactions on Intelligent Systems and Technology (TIST)  \textbf{3}(4),  1--21 (2012)

\bibitem{wang2017liar}
Wang, W.: {``Liar, liar pants on fire'': A } new benchmark dataset for fake news detection. arXiv preprint arXiv:1705.00648  (2017)

\bibitem{10.1145/3508398.3519358}
Zeng, V., Liu, X., Verma, R.M.: Does deception leave a content independent stylistic trace? In: In Proc. of ACM CODSAPY. p. 349–351 (2022)

\bibitem{zhang2018dri}
Zhang, W., Du, Y., Yoshida, T., Wang, Q.: {DRI-RCNN}: An approach to deceptive review identification using recurrent convolutional neural network. Information Processing \& Management  \textbf{54}(4),  576--592 (2018)

\bibitem{zhang2019dcword}
Zhang, W., Wang, Q., Li, X., Yoshida, T., Li, J.: Dcword: a novel deep learning approach to deceptive review identification by word vectors. Journal of Systems Science and Systems Engineering  \textbf{28},  731--746 (2019)

\bibitem{zhao2018towards}
Zhao, S., Xu, Z., Liu, L., Guo, M., Yun, J.: Towards accurate deceptive opinions detection based on word order-preserving cnn. Math. Problems in Engg.  \textbf{2018} (2018)

\bibitem{zubiaga2016learning}
Zubiaga, A., Liakata, M., Procter, R.: Learning reporting dynamics during breaking news for rumour detection in social media. arXiv preprint arXiv:1610.07363  (2016)

\end{thebibliography}

\end{document}